\documentclass{SCIS2026}
\usepackage{float}
\begin{document}
\ArticleType{RESEARCH PAPER}
\Year{2025}
\Month{January}
\Vol{68}
\No{1}
\DOI{}
\ArtNo{}
\ReceiveDate{}
\ReviseDate{}
\AcceptDate{}
\OnlineDate{}
\AuthorMark{}
\AuthorCitation{}

\title{E$^{3}$-Agent: An Executable and Evolving Agent for Resource Management of Edge Generative Inference}{Title for citation}

\author[1]{Rui Bao}{}
\author[2]{Yaping Sun}{}
\author[1]{Zhiyong Chen}{{zhiyongchen@sjtu.edu.cn}}
\author[1]{Feng Yang}{}
\author[1]{Meixia Tao}{{mxtao@sjtu.edu.cn}}
\author[3]{Nan Li}{}
\author[1]{Wenjun Zhang}{}


\address[1]{Cooperative Medianet Innovation Center, Shanghai Jiao Tong University, Shanghai 200240, China}
\address[2]{Department of Broadband Communication, Pengcheng
Laboratory, Shenzhen 518000, China}
\address[3]{ China Mobile Research Institute, Beijing 100053, China}


\abstract{Edge deployments of generative inference increasingly face two practical realities: per-device per-model performance is often unknown at deployment time, and it is non-stationary due to user-driven semantic events, background load, and device churn. Consequently, a resource manager that is tuned offline under a fixed regime can become brittle and expensive to maintain. This paper presents E$^{3}$-Agent, an executable and evolving agent for edge artificial intelligence generated content (AIGC) resource management. E$^{3}$-Agent separates a fast-path router that makes millisecond-level dispatch decisions from a slow-path, event-driven large language model (LLM) meta-controller that mitigates regime shifts through a small, explicit control surface exposed via a tool interface, including risk gating, router configuration, and rapid performance calibration. The agent learns online from execution feedback and continuously adapts to unknown and time-varying service-time mappings. We evaluate E$^{3}$-Agent in a discrete-event simulator driven by MLPerf-derived device--model measurement priors, covering cold-start warmup and three dynamic regimes: semantic dynamics, device churn, and hidden drift. Across the dynamic scenarios, E$^{3}$-Agent reduces average latency by 65\%--73\% compared to the best static baseline, stays within 7\%--10\% of an online full-information Oracle used for evaluation, and effectively suppresses stutter rate under semantic degradation.}

\keywords{agentic edge computing; LLM agentic system; online learning; resource management; performance drift}

\maketitle
\raggedbottom

\section{Introduction}
Edge deployments of generative inference are increasingly shaped by heterogeneous device pools and rapidly evolving model families. Compared to earlier mobile AI workloads, modern large language models (LLMs) and diffusion models (DMs) impose substantially higher compute and memory pressure and exhibit longer, more variable runtimes under interactive request patterns. Measurements of production LLM services report pronounced burstiness and wide variation in concurrency and response lengths, making it difficult to assume a single stationary operating regime~\cite{burstgpt2024}. At the same time, advances in serving stacks and memory systems continue to shift practical bottlenecks and operating points~\cite{vllm2023}. As a result, edge resource management must cope with per-device, per-model service times that are unknown at deployment time and non-stationary under changing workload and device conditions. Latency-sensitive immersive applications further tighten the operational constraints of edge systems and expose stronger coupling across communication and computing resources~\cite{sun2019communications}. Recent studies on multi-user interactive virtual reality highlight how motion-to-photon constraints amplify the need for responsive edge control under dynamic wireless and device conditions~\cite{xu2025wireless}. This need is consistent with recent 6G visions that identify AI-native communication, sustainable intelligent networking, and wireless-native big AI models as central research directions for future networks~\cite{cui2025overview6g,you2025sustainable6g,chen2025wirelessnative}.

A broad literature has studied task offloading and resource management in edge--cloud systems using heuristics and deep reinforcement learning~\cite{hortelano2023rlsurvey}. While learning-based approaches can approach optimality under well-specified models and sufficient interaction data, deployments often face a different challenge: sustaining performance under distribution shift. In realistic edge settings, task mixes, arrival intensities, and request sizes evolve over time, while underlying system conditions drift with device availability, contention, thermal throttling, and software upgrades. Recent mobile edge computing (MEC) studies explicitly observe that policies trained under a single approximately stationary regime may require substantial redesign and retraining when confronted with non-stationary streams or unfamiliar devices and tasks~\cite{khoshvaght2025zerosched}. The associated operational burden, including instrumentation, safety validation, and repeated redeployment, becomes a practical bottleneck for edge generative services, where models and workloads can change at a rapid cadence.

Edge AIGC further amplifies these challenges in three ways. First, interactive LLM applications increasingly appear as multi-step agentic workflows, in which application structure influences end-to-end latency and complicates per-request scheduling and placement decisions~\cite{parrot2024}. Second, diffusion-based generation remains compute-intensive, motivating edge--cloud collaboration to balance latency and cost, but also increasing sensitivity to inaccurate service-time assumptions~\cite{hybridsd2024}. Third, device heterogeneity at the edge spans embedded modules, consumer SoCs, and accelerator-equipped gateways, each subject to distinct power, thermal, and contention dynamics. Consequently, static routing or offloading rules tuned for a particular device mix or workload regime can degrade sharply as the device pool and request characteristics evolve. 

This paper argues for a resource manager that is explicitly designed as an executable closed loop. The key capability is to couple information acquisition, strategy adaptation, and online validation within the running system so that the policy remains effective as conditions change. This framing is motivated by two practical constraints in edge generative inference. First, service-time mappings are unknown at deployment and can shift under contention, thermal dynamics, and software evolution, so a one-time profiling pass cannot be assumed sufficient. Second, the control plane must remain auditable and safe under uncertainty, which calls for explicit control knobs, bounded intervention frequency, and measurable post-action validation. An executable closed loop makes these requirements first-class by enforcing that every adaptation is grounded in observed feedback and can be reverted or overridden when risk signals arise.

Recent progress in agentic AI suggests a viable implementation path: tool-using LLM agents can orchestrate telemetry, profiling, and configuration interfaces to automate operational workflows. In telecom, foundation-model-driven designs have been discussed as a unifying substrate for operational automation across heterogeneous interfaces and modalities~\cite{telecomfm2024}. In parallel, surveys of LLM-based network management emphasize intent interpretation and tool coordination, while also highlighting reliability and domain adaptation as key barriers to safe deployment~\cite{hong2025llmnetworksurvey}. Practical intent-based networking prototypes further show how an LLM agent can translate intent into executable control actions grounded by controller context and verified outcomes~\cite{adanza2025intentibn}. Beyond terrestrial networks, SCNOC-Agentic demonstrates an explicit closed-loop decomposition with intent refinement, workflow orchestration, and grounded retrieval and memory for realistic operation and control tasks~\cite{sun2025scnoc}. Multi-agent LLM situation awareness has also been explored for zero-trust space-air-ground integrated networks, demonstrating the potential of LLM agents for sensing and reasoning in complex 6G environments~\cite{cao2025llmsa}. In O-RAN settings, recent work stresses that multi-tool agency must be paired with safety mechanisms and anomaly-aware control to avoid operational instability~\cite{salama2025edgeagenticaiframework}. Complementary wireless studies compare RL, rule-based, and generative-AI policies for Wi-Fi slicing, and show that prompt-driven policies can remain competitive while offering operational adjustability under traffic changes~\cite{intelwifi}. More recent proposals further model each access point as an LLM agent that negotiates coordination strategies through dialogue grounded in memory and tool use under dynamic interference and topology~\cite{fan2025learningmultiaccesspointcoordination}. Tutorials on agentic AI for future communications summarize planners, tool interfaces, and memory modules as reusable building blocks for adaptive systems~\cite{jiang2025largeaimodelsagentic}. More generally, ReAct formalizes a tool-mediated agent pattern that interleaves reasoning with action execution, enabling closed-loop decision making with external interfaces~\cite{yao2023react}. SWE-agent further demonstrates that agent-computer interface design materially affects end-to-end performance on complex, multi-step tasks and encourages explicit, auditable tool use~\cite{SWEagent}. Outside networking, recent work on LLM-augmented hierarchical control shows a practical timescale separation where the LLM provides high-level guidance at a reduced frequency while a fast module handles frequent actions~\cite{li2025llmdrive}.

Compared with these prior systems, E$^{3}$-Agent addresses a narrower but operationally distinct problem: online resource management for edge generative inference when per-device service-time mappings are initially unknown and later become non-stationary. Existing LLM network-management agents mainly focus on intent translation, situation awareness, or network-operation workflow automation; LLM serving systems primarily optimize cluster-side serving under known infrastructure; and MEC offloading studies often assume a specified model or train policies under relatively fixed regimes. In contrast, E$^{3}$-Agent couples online performance estimation, event-driven meta-control, and a constrained executable tool interface so that adaptation is driven only by causal execution feedback and can be audited after each intervention.

Building on this insight, we present \textbf{E$^{3}$-Agent}, an executable and evolving agentic framework for edge generative inference. E$^{3}$-Agent is deployed on the edge server at the base station (BS) and positions the LLM as an event-driven meta-controller that adjusts a small, explicit control surface while leaving per-task dispatch to a lightweight fast-path router. The system learns online from execution feedback to estimate device--task service-time mappings, reacts to semantic and performance anomalies via risk gating, and rapidly recalibrates after regime shifts. This separation enables near-oracle performance with bounded control-plane overhead. The main contributions of this paper are summarized as follows:

\begin{itemize}
    \item \textbf{Executable closed-loop agentic framework.} We propose E$^{3}$-Agent, an end-to-end executable framework for edge generative inference resource management that couples information acquisition, strategy adaptation, and online validation in a running system. A key design choice is a principled separation between a millisecond-level fast path for per-task dispatch and an event-driven LLM meta-controller that operates over an explicit, auditable control surface.
    
    \item \textbf{Online learning and non-stationary adaptation.} We introduce an online learning loop that operates under initial uncertainty regarding device--task performance mappings and continuously updates an agent-side performance model from execution feedback. The same loop supports rapid mitigation under non-stationarity, including semantic events, device churn, and hidden performance drift, while enforcing safety-relevant routing constraints and keeping control-plane overhead bounded through event-driven triggering.
    
    \item \textbf{Evaluation with real-measurement priors.} Using MLPerf-derived device--model measurement priors as realistic profiles~\cite{mlperf}, we evaluate E$^{3}$-Agent in a discrete-event simulator under cold-start warmup and three dynamic regimes. Across the dynamic scenarios, E$^{3}$-Agent reduces average latency by about 65\%--73\% compared to the best static baseline while remaining within 7\%--10\% of an online full-information Oracle.
\end{itemize}

The remainder of this paper is organized as follows. Section~\ref{sec:problem} formalizes the edge generative inference setting and formulates the resource management problem. Section~\ref{sec:agent} presents the architecture and design of E$^{3}$-Agent, including its tool layer, state, and event-driven meta-control loop. Section~\ref{sec:experiments} evaluates E$^{3}$-Agent under heterogeneous device pools and non-stationary task regimes, and analyzes the impact of closed-loop adaptation. Section~\ref{sec:conclusion} concludes the paper and discusses future directions.

\section{System Model and Problem Formulation}
\label{sec:problem}

\begin{figure*}[t]
    \centering
    \includegraphics[width=1\textwidth]{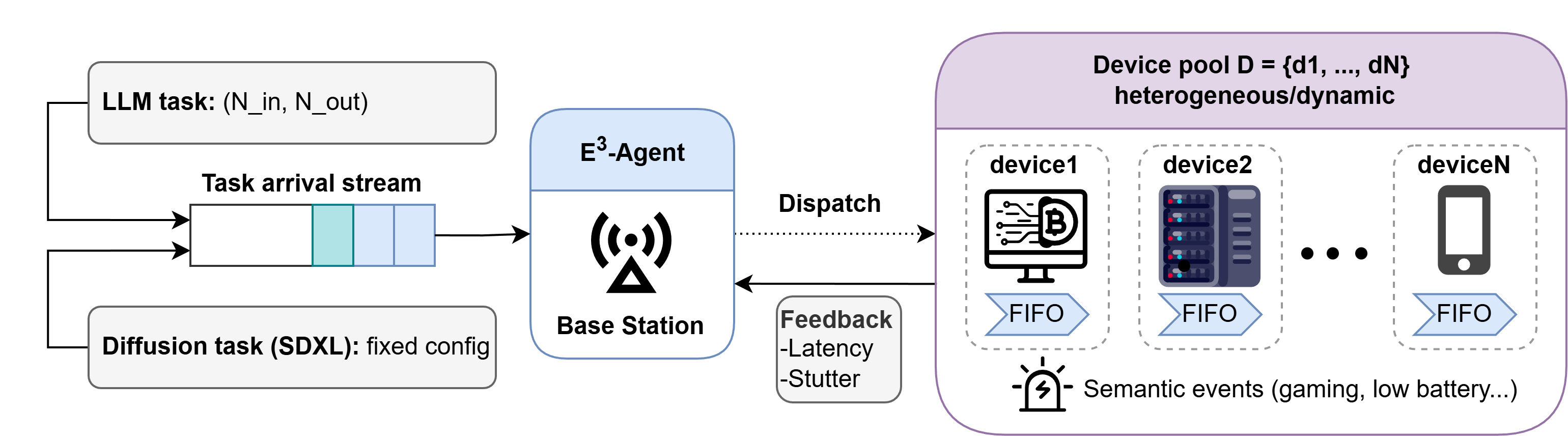}
    \caption{System model of E$^{3}$-Agent. A central controller routes a mixed stream of LLM and SDXL tasks to a heterogeneous device pool with FIFO queues and closes the loop using execution feedback, including latency and stutter under semantic events.}
    \label{fig:system_model}
\end{figure*}

As shown in Fig.~\ref{fig:system_model}, we consider an edge computing environment where a central controller, implemented as E$^{3}$-Agent at a base station, manages a stream of generative AI inference tasks for $N$ user devices via wireless networks. The system handles a heterogeneous arrival stream consisting of LLM tasks, characterized by input and output token counts $N_{in}$ and $N_{out}$, and Stable Diffusion tasks based on SDXL with fixed configurations.

These tasks are dispatched to a pool of $N$ heterogeneous and dynamic devices $\mathcal{D} = \{d_1, \dots, d_N\}$, ranging from high-performance workstations with discrete GPUs to mobile devices with integrated NPUs. Each device $d_i$ maintains a local FIFO queue for task execution. To achieve closed-loop control, E$^{3}$-Agent continuously monitors the execution status and receives feedback, including latency and stutter metrics. Crucially, the system must adapt to semantic events, such as background gaming and low battery, which cause non-stationary fluctuations in resource availability. These semantic events represent user-driven environmental shifts that are not directly observable but can significantly impact end-to-end performance.

\paragraph{Deployment and wireless-scope assumption.}
The controller is placed at the BS-side edge server and makes dispatch decisions after a task request has entered the edge service domain. In the present simulator, we do not separately model PHY/MAC scheduling, channel fading, or radio retransmissions. Instead, the measured service profile represents the device--model inference component used by the resource manager, while communication latency is treated as an exogenous component that is either common across compared policies or can be incorporated into deployment telemetry. Therefore, the conclusions of this paper concern closed-loop compute-side resource management under a given access condition, rather than joint wireless-channel and computing-resource optimization.

\begin{table*}[t]
    \centering
    \footnotesize
    \setlength{\tabcolsep}{5pt}
    \renewcommand{\arraystretch}{1.15}
    \caption{Information and timing model used by E$^{3}$-Agent. The agent and baselines use only causal observations; simulator ground truth is reserved for the Oracle evaluator.}
    \label{tab:info_model}
    \begin{tabular}{p{2.7cm} p{6.2cm} p{5.4cm}}
    \toprule
    Category & Available information & Timing and access rule \\
    \midrule
    Decision-time observables & Task type and task features, current available device set, queue states, active risk overrides, OPM estimates, sample counts, and event annotations already exposed to the controller & Available before dispatch of task $k$ \\
    Hidden variables & True time-varying service parameters $\theta_d(t)$, hidden degradation state $z_d(t)$, degradation magnitude, future arrivals, and future completions & Not accessible to E$^{3}$-Agent or baselines \\
    Delayed feedback & Completion latency, service-time component, LLM token lengths, stutter indicator in semantic dynamics, and residual statistics derived from completed tasks & Appended only after task completion and consumed causally by OPM \\
    Oracle-only information & Current ground-truth $\theta_d(t)$ and $z_d(t)$ for the feasible devices & Used only to form an online upper-bound baseline; it does not see future arrivals \\
    \bottomrule
    \end{tabular}
\end{table*}

\subsection{Workload and Unknown Mapping}
The workload consists of two primary generative task types: large language models, denoted LLMs, and diffusion models. The true service time $T^*$ of a task on device $d$ is governed by device-specific parameters that are initially unknown to the resource manager. This is because edge deployments face heterogeneous and evolving device pools, including newly joined devices, and reliable per-device per-model service-time profiles are often unavailable or outdated due to model and serving-stack evolution as well as limited profiling coverage. The mapping must therefore be learned online from execution feedback.

\textbf{LLM Tasks:} LLM tasks are characterized by input token length $N_{in}$ and output token length $N_{out}$. We model the service time using a standard TTFT/TPOT decomposition~\cite{zhong2024distserve}:
\begin{equation}
    T^*_{llm}(d) = \alpha_d(t) N_{in} + \beta_d(t) N_{out},
\end{equation}
where $t$ denotes the current online decision epoch, i.e., the index of the task-arrival and dispatch event. The coefficients $\alpha_d(t)$ and $\beta_d(t)$ have units of milliseconds per token and represent the time-to-first-token sensitivity and time-per-output-token, respectively.

This is an empirical linear approximation commonly used to decompose LLM latency into a prefill-dominated component and a decode-dominated component. Intuitively, TTFT is largely incurred by prefill, and under fixed model and device conditions it is approximately linear in input length. TPOT captures the average per-token time in the decode stage, so the total decode time is approximately $\beta N_{out}$. In our edge setting, $\alpha_d(t)$ and $\beta_d(t)$ are time-varying because the effective device condition changes over time, including semantic events, dynamic background load, thermal or power throttling, and device join or leave. We therefore estimate them online using recent execution feedback.

\textbf{Diffusion Tasks:} Modeled as a fixed-complexity task for a given configuration. We use Stable Diffusion XL, denoted SDXL, at $1024\times1024$ resolution~\cite{podell2024sdxl}. The service time is:
\begin{equation}
    T^*_{sd}(d) = \gamma_d(t).
\end{equation}

Crucially, the coefficients $\theta_d(t) = \{\alpha_d, \beta_d, \gamma_d\}$ are time-varying and hidden because device conditions change over time, including background user activities, contention, thermal or power throttling, and device join or leave. This leads to non-stationary service times even for the same device--task pair, so the system must estimate them online as $\hat{\theta}_d$.

\subsection{Semantic Events and Hidden Degradation}
We introduce a \textbf{semantic event-driven} degradation mechanism. The performance state of device $d$, denoted by $z_d(t)$, transitions between two regimes, $\mathcal{Z} = \{\text{Stable}, \text{Degraded}\}$, triggered by semantic events $E_t$ such as user gaming and low battery.
The causal chain is as follows. A semantic onset may be exposed to the controller as a coarse event or telemetry hint, but the actual degradation state and its magnitude remain hidden. When a device enters a degraded state, for example due to GPU resource contention caused by gaming, its effective service-time parameters can shift abruptly and nonlinearly. The router therefore cannot read the true degradation magnitude directly and must infer the performance impact from subsequent execution feedback. In addition to latency, such states can cause perceptible Quality-of-Experience (QoE) impairment if heavy generative tasks are dispatched to the affected device. We model this QoE risk using a binary stutter signal $S_k\in\{0,1\}$ for task $k$ dispatched to device $d$:
\begin{equation}
        S_k = \mathbb{I}(z_{d}(t) = \text{Degraded}).
\label{eq:stutter}
\end{equation}
This formulation treats stutter as an observable post-dispatch QoE indicator in semantic dynamics. In a deployment, the same signal can be obtained from application telemetry or device-side runtime logs; in our simulator it is recorded as an event annotation after the affected task is dispatched. The signal enables risk-aware routing and mitigation without introducing an additional hard constraint into the optimization objective.

\subsection{Problem Statement: Latency Minimization and Oracle Gap}
The goal is to design a policy $\pi$ that maps the current system state $s_t$, such as queue lengths, task features, and event history, to an assignment $a_t \in \mathcal{D}$.
Our objective is to minimize the average end-to-end latency over a horizon $H$:
\begin{equation}
    \min_\pi \frac{1}{H} \sum_{k=1}^H L_k(\pi),
\end{equation}
where $L_k(\pi)$ denotes the end-to-end latency of task $k$ under policy $\pi$, including queueing delay.
We also define an online full-information Oracle, which makes routing decisions using the ground-truth, time-varying parameters $\theta_d(t)$ and hidden device states $z_d(t)$ at the current time, and does not use future arrivals. This access is intentionally stronger than that of E$^{3}$-Agent and all practical baselines; the Oracle is not an implementable policy in our information model. We use this Oracle only as an upper-bound baseline to quantify how quickly an adaptive agent converges under unknown and non-stationary environments. We report performance using the Oracle gap:
\begin{equation}
    \frac{1}{H} \sum_{k=1}^H (L_k(\pi) - L_k(\text{Oracle})).
\end{equation}
In addition to latency, we report stutter defined in Eq.~\ref{eq:stutter} as a QoE risk signal in semantic dynamics. The agent should keep this signal low via risk-aware routing and event-driven mitigation, and reduce the Oracle gap quickly after each regime shift.

\section{E$^{3}$-Agent Architecture}
\label{sec:agent}

\subsection{Design Rationale}
E$^{3}$-Agent is built around a simple principle: separate millisecond-level routing from slow, semantic adaptation. Per-task dispatch must be fast and stable, operating at high frequency with low overhead, while the system still needs the ability to explore unknown mappings, detect regime shifts, and react to semantic events at a lower frequency with higher-level reasoning. Therefore, we position the LLM as a \textbf{meta-controller} that orchestrates a small set of executable control actions, rather than issuing per-task routing decisions directly.

This design is motivated by three constraints that arise in edge generative inference. First, the system must respect timescale separation: routing decisions occur at task arrival and should be executed deterministically with bounded overhead, whereas semantic interpretation, such as detecting the onset of gaming, and drift reasoning can be event-driven. Second, the control surface must remain minimal and explicit so that the system is implementable and auditable; the LLM is restricted to a small set of high-impact control parameters that map directly to executable system actions, including risk overrides and calibration intensity. Third, the system must satisfy a non-leakage and reproducibility contract: the agent must not access simulator ground truth, and all learning and adaptation are driven only by past execution feedback under a deterministic experimental configuration.

\subsection{Architecture Overview}
\begin{figure*}[t]
    \centering
    \includegraphics[width=\textwidth]{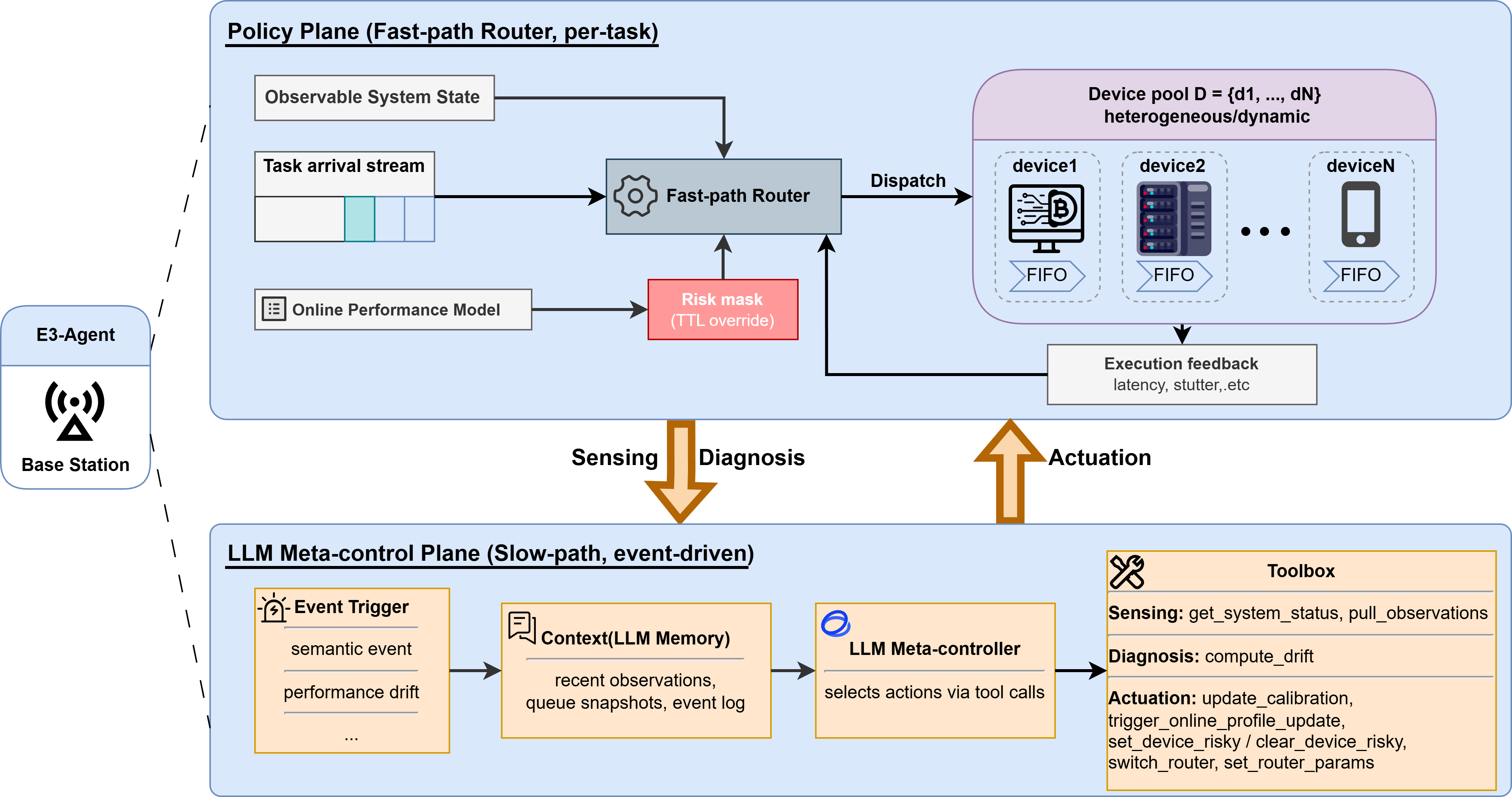}
    \caption{E$^{3}$-Agent architecture. The policy plane runs a fast-path router for per-task dispatch using queue states, OPM-predicted service times, and a risk mask. The meta-control plane is slow-path and event-driven: it maintains context as LLM memory and uses a small tool interface for sensing, diagnosis, and actuation. Sensing and diagnosis query system status and recent observations and quantify model--reality mismatch, including drift and residual patterns, to decide when and how to actuate. Actuation updates the OPM, risk overrides, and router configuration, and the policy plane consumes these updates to shape subsequent dispatch decisions as execution feedback closes the loop.}
    \label{fig:arch}
\end{figure*}

Fig.~\ref{fig:arch} illustrates the architecture of the proposed E$^{3}$-Agent. The architecture consists of two coupled loops: a \textbf{fast data loop} that schedules each task using a lightweight router, and a \textbf{slow control loop} that is triggered by events or anomalies and adjusts a small set of control knobs.

\textbf{Fast data loop.} A task stream arrives online with two generative task types: LLM and SDXL. For each incoming task, the router selects a device based on the current queue states, the agent-side online performance estimates, and a risk mask. Each device is modeled as a FIFO queue with a single server. After execution, the system captures task-level feedback, including end-to-end latency and the service-time component excluding queueing delay, as well as semantic stutter indicators in Eq.~\ref{eq:stutter} when applicable.

\textbf{Slow control loop.} Fig.~\ref{fig:arch} organizes this two-loop design into two interacting planes. The policy plane runs a fast-path router on every task arrival. The meta-control plane is slow-path and event-driven. The Online Performance Model, OPM, together with device profiles, historical observations, and an event log, provides shared state consumed by both planes. When triggers fire, the LLM meta-controller queries recent telemetry for sensing and diagnosis, then updates risk overrides, calibration settings, and router configuration via tools. The policy plane consumes these updates immediately in subsequent routing decisions as new execution feedback arrives.

\subsection{Policy Plane: Fast-path Routing}
As shown in Fig.~\ref{fig:arch}, the policy plane runs for every arriving task and must remain lightweight. Conceptually, it can be any executable router, whether heuristic or learned, that maps the current observable system state to a device assignment. In E$^{3}$-Agent, the router is driven by three classes of signals: queue and backlog information, predicted service time from the OPM, and a risk signal that encodes safety and QoE constraints.

To keep the discussion concrete, we describe a generic Shortest Expected Completion Time (SECT)-style scoring form that captures the essential ingredients. SECT estimates each device's expected completion time as the sum of queueing delay and the predicted service time, and selects the device with the smallest estimate:
\begin{equation}
    d^* = \arg\min_{d} \left( Q_d(t) + \hat{T}_d(k) - c u_d(k) + M_d(t) \right),
\end{equation}
where $Q_d(t)$ summarizes the current backlog at device $d$, $\hat{T}_d(k)$ is the predicted service time, $u_d(k)$ is epistemic uncertainty, and $M_d(t)$ is a risk penalty. The key is not the exact functional form but the division of responsibilities: $\hat{T}$ and $u$ come from the OPM and its profile cache, while $M$ comes from risk gating configured by the meta-control plane. This allows the fast-path to remain simple while still adapting to unknown mappings and regime shifts.

The feasible action set contains only devices that are currently available. When a risk override is active, E$^{3}$-Agent applies hard avoidance if at least one non-risky device remains. If all available devices are risky, the router does not deadlock; it retains the available devices and selects the least-bad assignment according to the same SECT score with risk penalties. This fallback represents a route-anyway policy under degraded service availability, leaving admission control or task dropping to an outer service-level policy.

\subsection{Model and Memory: Online Performance Model}
The OPM maintains per-device performance estimates without accessing simulator ground truth. This design constraint ensures that the agent learns the system mapping exclusively from execution feedback. To achieve this, the OPM utilizes the measured service time of completed tasks, defined as the total execution duration minus queuing delay, to isolate inherent device--model performance from transient noise.

Each execution record captures the device ID, task type, end-to-end latency, and the service-time component. For LLM tasks, the system further logs the input and output token lengths $N_{in}$ and $N_{out}$. For semantic dynamics, it records the stutter indicator in Eq.~\ref{eq:stutter}. The OPM operates on a causal basis, consuming only records available at or before the current timestamp. Feedback is therefore asynchronous at the task-completion granularity: an unfinished task cannot update the OPM, and delayed or missing feedback simply leaves the current estimate and uncertainty state unchanged until a valid record arrives.

For LLM tasks, we fit Eq.~(1) on a recent observation window using least squares for each device $d$:
\begin{equation}
    \min_{\hat{\alpha}_d,\hat{\beta}_d}\ \sum_{i \in W_d} \left(T^{(i)} - (\hat{\alpha}_d N_{in}^{(i)} + \hat{\beta}_d N_{out}^{(i)})\right)^2,
\end{equation}
and clip coefficients to be non-negative. This estimator is intentionally simple: it is data-efficient under short horizons and stable under small windows. The fitted coefficients $\hat{\alpha}_d$ and $\hat{\beta}_d$ are stored in the agent-side profile cache and updated online during warmup and after detected drift. In our experiments, token lengths are drawn from discrete bins described in Section~\ref{sec:experiments}, which provides controlled diversity while still requiring the agent to learn a non-trivial mapping.

For diffusion tasks, specifically SDXL, we treat each task as fixed complexity and calibrate a scalar $\hat{\gamma}_d$ based on recent observed service times. At initialization, the OPM is seeded with MLPerf-derived profiling priors; these priors are treated as imperfect and are progressively corrected by online observations.

To guide exploration and detect shifts, the OPM maintains a lightweight notion of epistemic uncertainty based on recent residual statistics, such as the variance of $T^{(i)}-\hat{T}(i)$ within a sliding window, and on sample scarcity reflected by small window sizes. Elevated uncertainty indicates either cold start due to insufficient observations or a recent regime shift due to model mismatch, motivating exploration in the policy plane or rapid recalibration in the meta-control plane. The linear TTFT/TPOT model is intentionally simple and may be mismatched under batching effects, thermal transitions, or software-stack changes. Such mismatch affects routing through biased $\hat{T}$ values, but it also produces residual growth that triggers recalibration; severe or persistent mismatch is therefore handled as a drift signal rather than as a hidden assumption that the router can ignore.

\subsection{Meta-control Plane: Event-driven LLM Controller}
The meta-control plane is invoked only when necessary. This is essential for both cost and stability: calling the LLM at a fixed interval can lead to unnecessary overhead and noisy actuation. We adopt an event-driven triggering design aligned with our problem definition of non-stationarity.
In our experiments, LLM calls are triggered by three classes of signals: semantic or drift events, where a device enters or leaves an abnormal state; residual alarms induced by sustained OPM mismatch; and a small number of warmup intervention points that accelerate convergence in unknown environments. Repeated triggers are suppressed by a signature-based cooldown, and non-event invocations obey a minimum task-count gap, so the meta-controller behaves as an exception handler rather than a periodic controller. The concrete default values are summarized in Table~\ref{tab:operational_defaults}.

Given the current system status and recent observations, the meta-controller acts over a limited control surface. In particular, it may apply or clear a risk override with a task-count time-to-live, TTL, to enforce strict stutter avoidance, and it may request rapid calibration by reducing the minimum samples required for an OPM update during warmup or after a detected regime shift. To keep actuation executable and auditable, we expose a small tool interface that maps directly to system actions for sensing, diagnosis, and actuation. Sensing retrieves recent outcomes and queue states, diagnosis quantifies model--reality mismatch from historical observations, and actuation updates risk gating, calibration intensity, and router configuration. Importantly, these tools do not reveal ground-truth service times; they only expose telemetry and historical observations that would be available in a deployed system.

\begin{table*}[t]
    \centering
    \scriptsize
    \setlength{\tabcolsep}{4pt}
    \renewcommand{\arraystretch}{1.15}
    \caption{Executable tool interface and constrained meta-controller action space. The LLM can only act through these tools and cannot directly dispatch individual tasks or inspect simulator ground truth.}
    \label{tab:tool_interface}
    \begin{tabular}{p{2.8cm} p{4.6cm} p{4.2cm} p{3.2cm}}
    \toprule
    Tool & Inputs & Output and effect & Constraint and audit record \\
    \midrule
    \texttt{get\_system\_status} & none & Queue length, utilization, current time, and exposed semantic state & Sensing only; logs query time \\
    \texttt{pull\_observations} & Window length and result limit & Recent completed-task summaries and stutter count & Uses completed observations only \\
    \texttt{compute\_drift} & Device ID or name, model, window length & Observed-to-predicted latency ratio and sample count & No ground-truth service time is returned \\
    \texttt{update\_calibration} & Device, model, calibration factor & Updates the OPM profile calibration factor & Logs old and new factors \\
    \texttt{switch\_router} & Router in \{\texttt{sect}, \texttt{explore\_risk}\} & Changes the fast-path scoring policy & Router choices are enumerated \\
    \texttt{set\_router\_params} & Exploration weight and risk penalty & Updates scoring parameters in milliseconds & Bounded scalar control surface \\
    \texttt{trigger\_online\_}\allowbreak\texttt{profile\_update} & Window length and minimum samples & Refits OPM from recent observations & Causal fitting from completed tasks \\
    \texttt{set\_device\_risky} and \texttt{clear\_device\_risky} & Device ID and optional TTL & Applies or clears risk override & Time-bounded; consumed by fast-path risk mask \\
    \bottomrule
    \end{tabular}
\end{table*}

Event-driven control is particularly important in edge settings where both workload and system conditions are bursty. Periodic LLM calls can waste budget during stable periods, amplify actuation noise by reacting to short-term variance, and introduce non-negligible control-plane latency. In contrast, event-driven triggers align control effort with regime shifts; in semantic dynamics, reacting promptly to semantic onset is essential to prevent repeated QoE impairments, while subsequent LLM invocations refine TTL choice and calibration intensity when necessary.

\subsection{Executable Closed-loop}
E$^{3}$-Agent implements an \textbf{executable} closed-loop by tightly coupling monitoring, diagnosis, actuation, and online model updates. Rather than relying on implicit ``prompt-only'' reasoning, each step maps to an explicit system state and an executable action.
The loop operates as follows. The system continuously collects task-level execution feedback, including service times and residuals, together with event annotations such as semantic state changes, join or leave, and drift indicators. The OPM computes uncertainty and drift statistics from recent observations; in semantic dynamics, semantic onset is exposed as an event signal, while the magnitude of degradation is inferred from subsequent feedback. When a trigger fires, the meta-controller selects from a small action set, primarily risk overrides with a TTL and calibration-intensity adjustments. The fast-path router then schedules every arriving task using the updated OPM estimates and the current risk mask, ensuring stable, low-overhead behavior even when the meta-controller is idle. Finally, new observations are appended to memory and the OPM updates its parameters, reducing uncertainty over time and improving routing decisions across repeated regimes.
In this paper, auditable means that the system records the inputs used for sensing, tool calls and arguments, tool return values, profile and calibration changes, risk-mask updates, and subsequent validation metrics such as residuals, latency, and stutter. The intended guarantee is not formal verification of optimality or safety, but a reproducible and bounded control surface: each adaptation can be traced to causal telemetry, checked against post-action outcomes, and rate-limited or cleared when the corresponding event disappears.

\subsection{Instantiation in Our Prototype}
The above architecture is intentionally modular; our experiments instantiate it in a minimal yet fully executable form. On the fast path, we use a SECT-style router augmented with an exploration signal and a risk gate: it routes tasks using queue and backlog information and OPM-predicted service times, while avoiding devices deemed risky under semantic events.

On the slow path, the OPM is updated online from recent execution feedback. For LLM tasks, it fits the token-to-latency relationship on a recent window, and for SDXL it calibrates a scalar service-time estimate. The meta-controller is event-driven: semantic onsets and offsets and drift alarms trigger risk gating and rapid recalibration, while stable periods incur no additional control-plane overhead. These choices match the implementation evaluated in Section~\ref{sec:experiments}.
The instantiated loop enforces safety and bounded overhead by construction. When any safe device exists, tasks are never dispatched to devices under an active risk override, implementing hard avoidance that prevents repeated QoE violations even if the OPM is temporarily inaccurate. Risk overrides are time-bounded via TTL and are cleared on semantic offset, allowing the system to re-admit recovered devices without manual intervention. Device churn dynamics are handled by state hygiene rules: unavailable devices are removed from the feasible action set, their stale queue state is not used for routing while offline, and their residual and profile statistics are refreshed from new observations after return before they are treated as high-confidence estimates. Triggers are rate-limited by cooldowns and minimum call gaps to prevent oscillations. From a systems standpoint, the fast-path router runs in $O(|\mathcal{D}|)$ per task by scoring each available device; OPM updates are performed over a bounded recent window; and the meta-controller is invoked only on triggers, so control-plane cost scales with regime shifts rather than the total task count. These design choices align with our evaluation: warmup isolates exploration and rapid calibration under unknown mappings, while the three dynamic regimes stress semantic risk control, robustness under device churn, and residual-driven recalibration under hidden drift.

\section{Experiments}
\label{sec:experiments}
\begin{table}[t]
    \centering
    \setlength{\tabcolsep}{12pt} 
    \renewcommand{\arraystretch}{1.2}
    \caption{MLPerf-derived workload configuration and deterministic device pool.}
    \label{tab:mlperf_workload_device}
    \begin{tabular}{p{1.5cm} p{3.5cm} p{6.5cm}} 
    \toprule
    \multicolumn{3}{l}{\textbf{Workload}} \\
    \midrule
    LLM  & \texttt{llama3.1-8b-edge} & Token bins: $N_{in}\in\{256,512,1024\}$, $N_{out}\in\{32,64,128\}$ \\
    SDXL & \texttt{stable-diffusion-xl} & Config: $1024\times1024$, 20 steps \\
    \midrule
    \multicolumn{3}{l}{\textbf{Device Pool}} \\
    \midrule
    LLM\#0 & NVIDIA Jetson Thor & SingleStream; fp4 \\
    LLM\#1 & Intel Arc Pro B60 & SingleStream; UINT4 \\
    SD\#0  & ThinkEdge SE100 & SingleStream; fp8 \\
    SD\#1  & NVIDIA Jetson Orin & SingleStream; fp8 \\
    \bottomrule
    \end{tabular}
    \end{table}

\subsection{Experimental Setup}
\begin{table*}[t]
    \centering
    \footnotesize
    \setlength{\tabcolsep}{5pt}
    \renewcommand{\arraystretch}{1.15}
    \caption{Operational defaults used in the reported experiments. These values are fixed unless the corresponding ablation explicitly varies them.}
    \label{tab:operational_defaults}
    \begin{tabular}{p{3.5cm} p{4.4cm} p{6.2cm}}
    \toprule
    Item & Default value & Role and rationale \\
    \midrule
    Task horizon and arrival & 300 tasks, $\lambda=0.5$ tasks per second, fixed inter-arrival interval & Reproducible convergence and dynamic adaptation analysis \\
    Task mixture & Alternating LLM and SDXL; LLM bins $\{256,512,1024\}\times\{32,64,128\}$ & Controlled diversity for online service-time fitting \\
    Warmup budgets & 0, 30, 100 tasks & Measures cold-start exploration and calibration effect \\
    Warmup LLM intervention points & $\{\min(10,W), W\}$ for warmup budget $W$ & Bounded early meta-control during unknown mapping \\
    OPM update during warmup & Every 5 tasks; recent-window fitting with minimum sample count 1 & Converts exploration samples into usable estimates quickly \\
    Trigger suppression & Minimum non-event LLM gap 20 tasks; anomaly cooldown 20 tasks & Suppresses repeated triggers and oscillation \\
    LLM tool rounds & At most 2 tool-use rounds per invocation & Bounds meta-control latency and cost \\
    Risk gate defaults & \texttt{risk\_penalty\_ms}=60000; risk TTL tool default 50 tasks & Enforces strong but time-bounded avoidance \\
    Exploration default & \texttt{explore\_weight\_ms}=2000 & Encourages sampling under sparse observations \\
    Drift defaults & Drift threshold 1.3, window 60 s, calibration smoothing 0.3 & Detects sustained residual mismatch and updates calibration \\
    \bottomrule
    \end{tabular}
\end{table*}

\paragraph{Simulator and queueing model.}
We implement a discrete-event simulator following Section~\ref{sec:problem}. Each device is modeled as a single-server queue without concurrency or micro-batching. Tasks are dispatched online and executed in FIFO order on each device. The simulator returns per-task completion latency and event annotations for semantic dynamics.

\paragraph{Dataset construction from MLPerf.}
To ensure realism, we construct a profiling dataset from standardized MLPerf Inference submissions rather than hand-tuned synthetic service times. Specifically, we use a curated JSONL file, \texttt{edge\_mobile\_generative.jsonl}, that filters MLPerf results to two generative benchmarks: \texttt{llama3.1-8b-edge} for LLM tasks and \texttt{stable-diffusion-xl} for diffusion tasks. Since edge generative services are latency-sensitive and interactive, we primarily use \textbf{SingleStream} measurements. For LLM profiles, we extract \texttt{ttft\_ms\_p99} and \texttt{tpot\_ms\_p99} and convert them into the linear prior in Eq.~(1). We set $\alpha=\texttt{ttft\_p99}/1024$ and interpret it in milliseconds per token, and we set $\beta=\texttt{tpot\_p99}$, also in milliseconds per token. For diffusion profiles, we use \texttt{latency\_ms\_p99} for SDXL at a fixed configuration with \texttt{image\_size}=1024 and \texttt{steps}=20.

\paragraph{Workload generation.}
We consider a two-task mixture that reflects modern edge generative inference. The first is LLM inference, parameterized by input and output token lengths $N_{in}$ and $N_{out}$. The second is diffusion inference with SDXL at fixed complexity. For LLM tasks, token lengths are sampled from discrete bins to enable repeatable convergence analysis under controlled diversity. Concretely, we sample discrete bins with a fixed mapping in the simulator: input bins map to $\{256,512,1024\}$ tokens and output bins map to $\{32,64,128\}$ tokens. Tasks arrive at a fixed interval of $1/\lambda$ with $\lambda=0.5$ tasks per second and alternate between LLM and SDXL unless otherwise stated. This deterministic stream is used to make convergence and non-stationary adaptation directly reproducible; the controller itself does not depend on a Poisson or i.i.d. arrival assumption.

\paragraph{Device pool.}
The device pool is derived from MLPerf Inference results filtered for edge and mobile generative models. These benchmark-derived measurements provide offline priors about device--model performance, which E$^{3}$-Agent treats as imperfect and continuously calibrates online. To avoid randomness in the evaluation setup, we use a deterministic heterogeneous device set shown in Table~\ref{tab:mlperf_workload_device}, selected from \texttt{edge\_mobile\_generative.jsonl}. The LLM-capable devices use 4-bit weights, fp4 or uint4 depending on the MLPerf submission, while SDXL devices follow the MLPerf SDXL configuration.

\paragraph{Execution feedback.}
For each executed task, the system records the realized latency. For LLM tasks, it also records the realized input and output token lengths, $N_{in}$ and $N_{out}$, enabling online fitting of the linear service-time model in Eq.~(1). For semantic dynamics, the system additionally records a binary stutter indicator in Eq.~\ref{eq:stutter} if a task is assigned to a device in a user-interactive or degraded state.


\paragraph{Dynamic regimes.}
We evaluate both cold-start learning, referred to as warmup, and non-stationary adaptation. In warmup, the system starts with an unknown device--task service-time mapping; we vary the warmup budget among 0, 30, and 100 tasks, during which E$^{3}$-Agent is encouraged to explore and quickly calibrate its online performance model before entering a steady routing regime. For dynamic environments, we use deterministic event plans over a 300-task horizon. Semantic dynamics contains five semantic windows, game, video call, low battery, system update, and overheating, which induce hidden slowdowns and stutter risk. Device churn changes the feasible device set through repeated device departures and returns. Hidden drift applies a step slowdown to one LLM device--model pair and later restores it, without exposing the drift factor to the agent.

\paragraph{Methods and baselines.}
We compare E$^{3}$-Agent against Fixed-Heuristic, a static SECT-style heuristic driven only by offline priors without online calibration or semantic risk control, and RoundRobin, a topology-agnostic baseline that cycles through available devices. We also report an online full-information Oracle that uses the ground-truth time-varying service times and hidden states at the current time, without using future arrivals, as a performance upper bound for latency.

\paragraph{Metrics.}
We report average end-to-end latency, measured in milliseconds, and the relative gap to Oracle, computed as $(\text{avg}/\text{oracle}-1)\times 100\%$ for each scenario. In semantic dynamics, we additionally report the stutter rate, defined as the fraction of tasks that trigger the stutter indicator in Eq.~\ref{eq:stutter}, reflecting QoE risk under user-interactive degradation.
For trajectory plots, we show sparsely sampled raw values every 5 tasks and a moving-average trend line, MA20, computed over a 20-task window.

\subsection{Results}

\begin{table*}[t]
    \centering
    \setlength{\tabcolsep}{3pt}
    \renewcommand{\arraystretch}{1.15}
    \caption{Overall results for Exp-1 Warmup and Exp-2 Dynamic. Lower is better. vs\_oracle in percent is computed as $((avg/oracle)-1) \times 100\%$. The Oracle column serves as the baseline with 0.00\%.}
    \label{tab:main_overall}
    \begin{tabular}{llrrrrrrr}
    \toprule
    & & \multicolumn{2}{c}{Fixed-Heuristic} & \multicolumn{2}{c}{RoundRobin} & \multicolumn{2}{c}{E$^{3}$-Agent} & \multicolumn{1}{c}{Oracle} \\
    \cmidrule(lr){3-4}\cmidrule(lr){5-6}\cmidrule(lr){7-8}\cmidrule(lr){9-9}
    \multirow{2}{*}{Block} & \multirow{2}{*}{Setting} & avg\_latency & vs\_oracle & avg\_latency & vs\_oracle & avg\_latency & vs\_oracle & avg\_latency \\
    & & ms & \% & ms & \% & ms & \% & ms \\
    \midrule
    \multirow{3}{*}{Warmup} & warmup=0   & 19475.24 & +288.04\% & 18500.67 & +268.62\% & 14363.78 & +186.19\% & 5018.93 \\
    & warmup=30  & 19475.24 & +288.04\% & 18500.67 & +268.62\% & 8052.86  & +60.45\%  & 5018.93 \\
    & warmup=100 & 19475.24 & +288.04\% & 18500.67 & +268.62\% & 6935.98  & +38.20\%  & 5018.93 \\
    \midrule
    \multirow{3}{*}{Dynamic} & semantic     & 19331.74 & +292.01\% & 19314.10 & +291.65\% & 5277.10  & +7.01\%   & 4931.43 \\
    & churn       & 16232.12 & +216.22\% & 17323.07 & +237.47\% & 5652.99  & +10.13\%  & 5133.23 \\
    & drift       & 16602.12 & +244.23\% & 18285.96 & +279.14\% & 5255.70  & +8.97\%   & 4823.00 \\
    \bottomrule
    \end{tabular}
\end{table*}

\begin{table}[t]
    \centering
    \footnotesize
    \setlength{\tabcolsep}{5pt}
    \renewcommand{\arraystretch}{1.15}
    \caption{QoE and meta-control overhead from the reported dynamic runs. LLM calls and tool calls are counted only for E$^{3}$-Agent.}
    \label{tab:qoe_overhead}
    \begin{tabular}{lrrrr}
    \toprule
    Scenario & E$^{3}$ stutter & Best static stutter & LLM calls & Tool calls \\
    \midrule
    Semantic & 0.00\% & 10.00\% & 11 & 33 \\
    Device churn & 0.00\% & 0.00\% & 2 & 6 \\
    Hidden drift & 0.00\% & 0.00\% & 4 & 16 \\
    \bottomrule
    \end{tabular}
\end{table}

\paragraph{Summary of results.}
Table~\ref{tab:main_overall} summarizes Exp-1 Warmup and Exp-2 Dynamic. We report average end-to-end latency and the relative gap to the online full-information Oracle. Warmup isolates the cold-start phase where device--task mappings are initially unknown and must be learned from feedback. Dynamic regimes stress adaptation under semantic degradation, device churn, and hidden drift, where the agent must maintain low latency while tracking time-varying performance conditions.

\begin{figure}[H]
    \centering
    \includegraphics[width=0.7\textwidth]{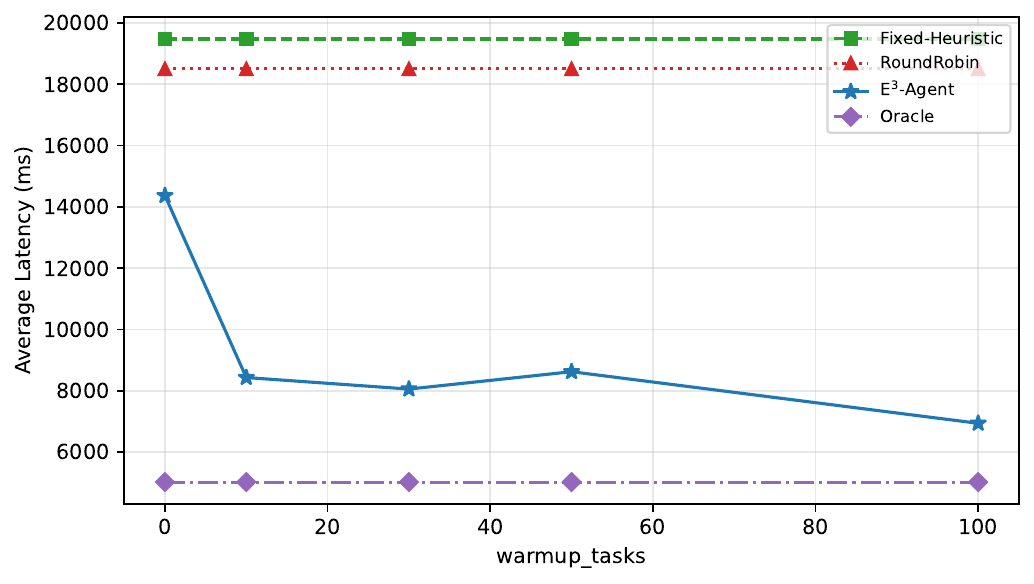}
    \caption{Warmup ablation: average latency curves under different warmup budgets using raw samples every 5 tasks with MA20 smoothing. MA20 is a 20-task moving average.}
    \label{fig:warmup_latency_curve}
\end{figure}

\paragraph{Warmup in unknown environments.}
This experiment evaluates whether E$^{3}$-Agent can reduce cold-start cost and converge toward Oracle when the device--task service-time mapping is initially unknown, as is common when deploying to a new device pool or after a model or stack update. We vary the warmup budget among 0, 30, and 100 tasks. During warmup, E$^{3}$-Agent prioritizes information acquisition and fast online calibration of Eq.~(1) so that the fast-path router becomes reliable in steady operation. Fig.~\ref{fig:warmup_latency_curve} shows a clear warmup-to-convergence trend: the gap to Oracle decreases from +186.19\% at warmup=0 to +60.45\% at warmup=30, and further to +38.20\% at warmup=100. These results indicate that early exploration combined with rapid online calibration materially accelerates convergence under unknown mappings, converting an initially uncertain environment into an effectively controllable regime for low-overhead routing.
Table~\ref{tab:main_overall} further shows that the improvement is driven by a sharp reduction in average latency for E$^{3}$-Agent as warmup increases, from 14363.78 ms at warmup=0 to 8052.86 ms at warmup=30 and 6935.98 ms at warmup=100. In contrast, Fixed-Heuristic and RoundRobin remain near 19 s across warmup budgets because they lack a mechanism to learn the mapping and therefore cannot correct mismatched priors. The remaining gap at warmup=100 reflects the finite-horizon nature of warmup in a heterogeneous pool, where limited samples per device and token bin constrain the precision of online estimates.

\begin{figure}[H]
    \centering
    \begin{minipage}[t]{0.49\textwidth}
        \centering
        \includegraphics[width=\textwidth]{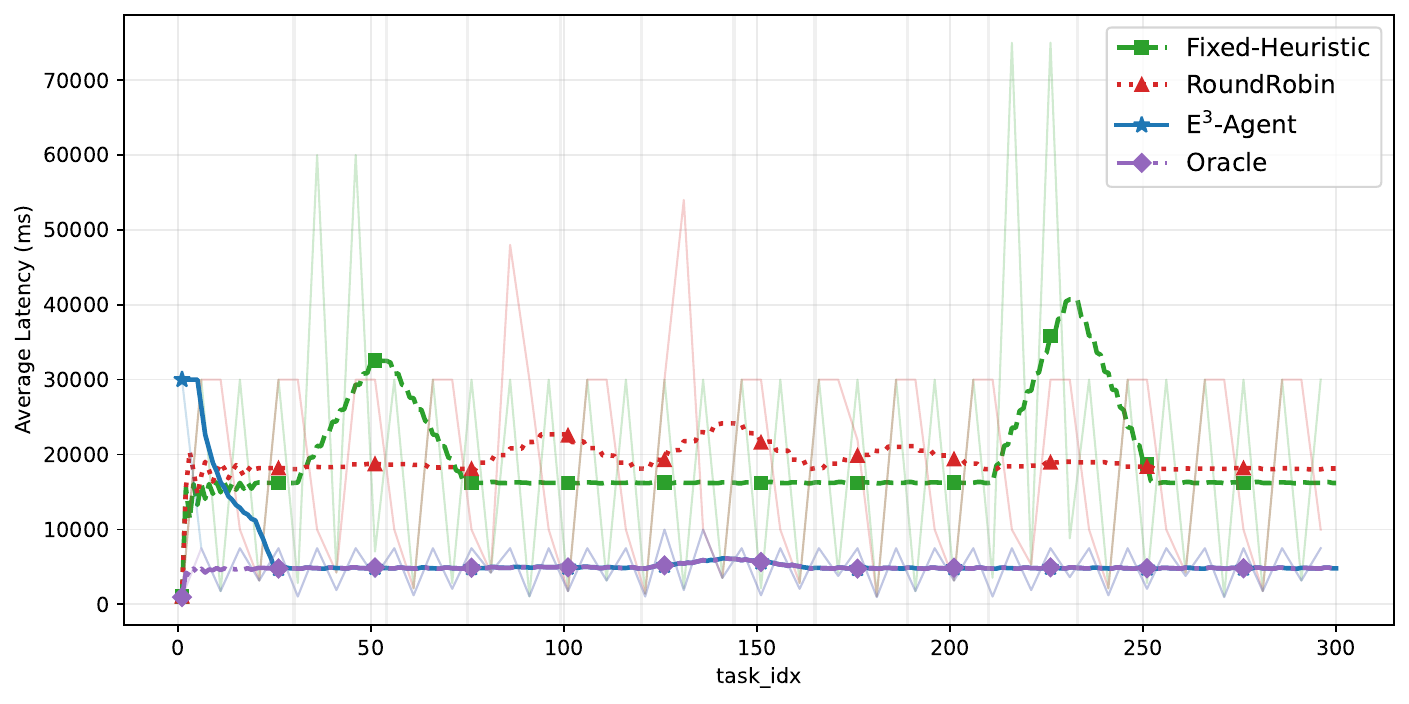}
        \par\small (a) Semantic dynamics (latency)
    \end{minipage}
    \hfill
    \begin{minipage}[t]{0.49\textwidth}
        \centering
        \includegraphics[width=\textwidth]{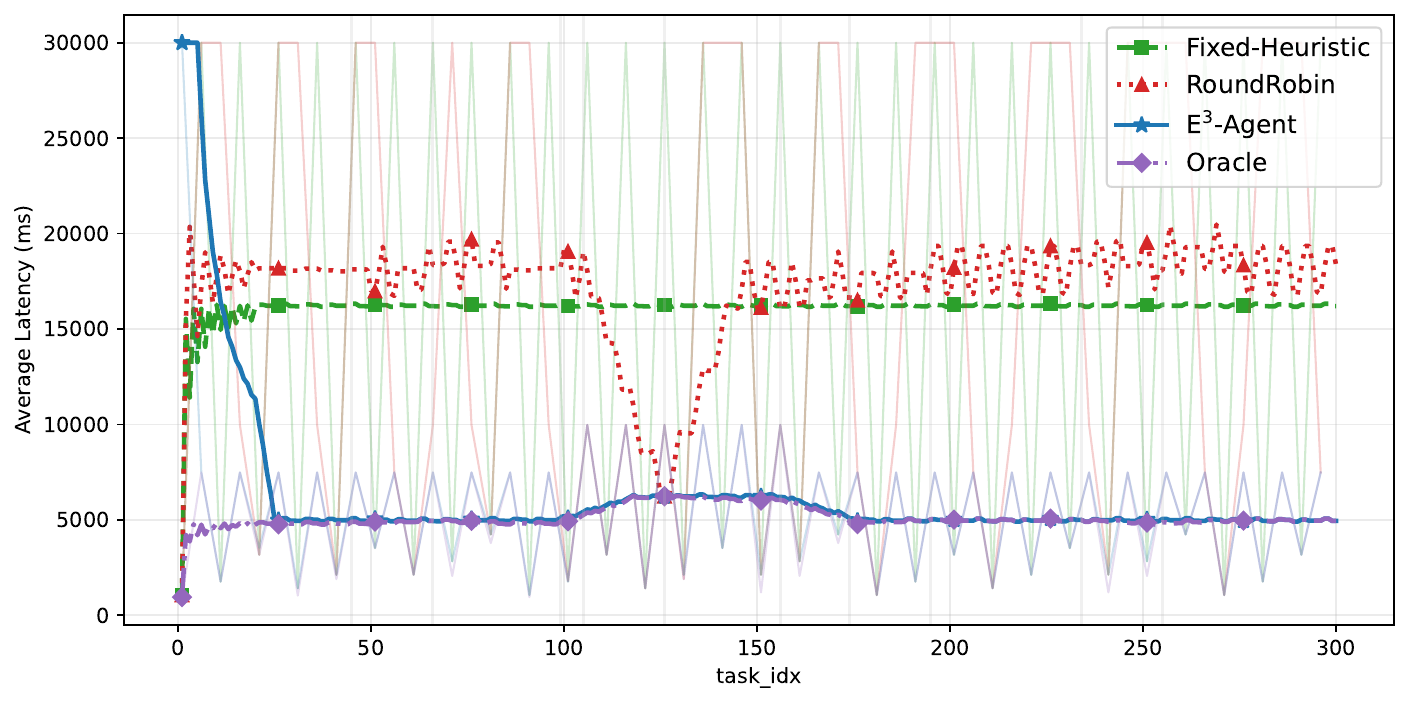}
        \par\small (b) Device churn dynamics (latency)
    \end{minipage}

    \vspace{0.6em}

    \begin{minipage}[t]{0.49\textwidth}
        \centering
        \includegraphics[width=\textwidth]{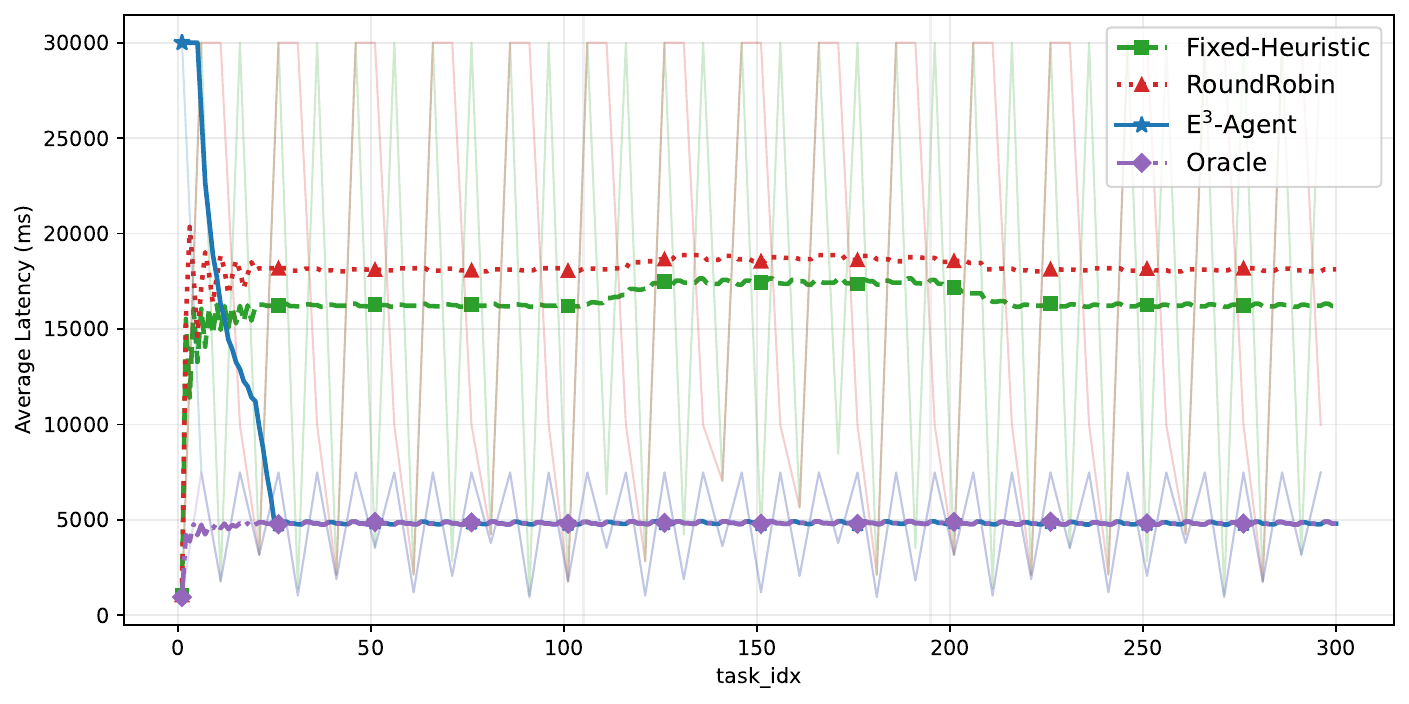}
        \par\small (c) Hidden drift (latency)
    \end{minipage}
    \hfill
    \begin{minipage}[t]{0.49\textwidth}
        \centering
        \includegraphics[width=\textwidth]{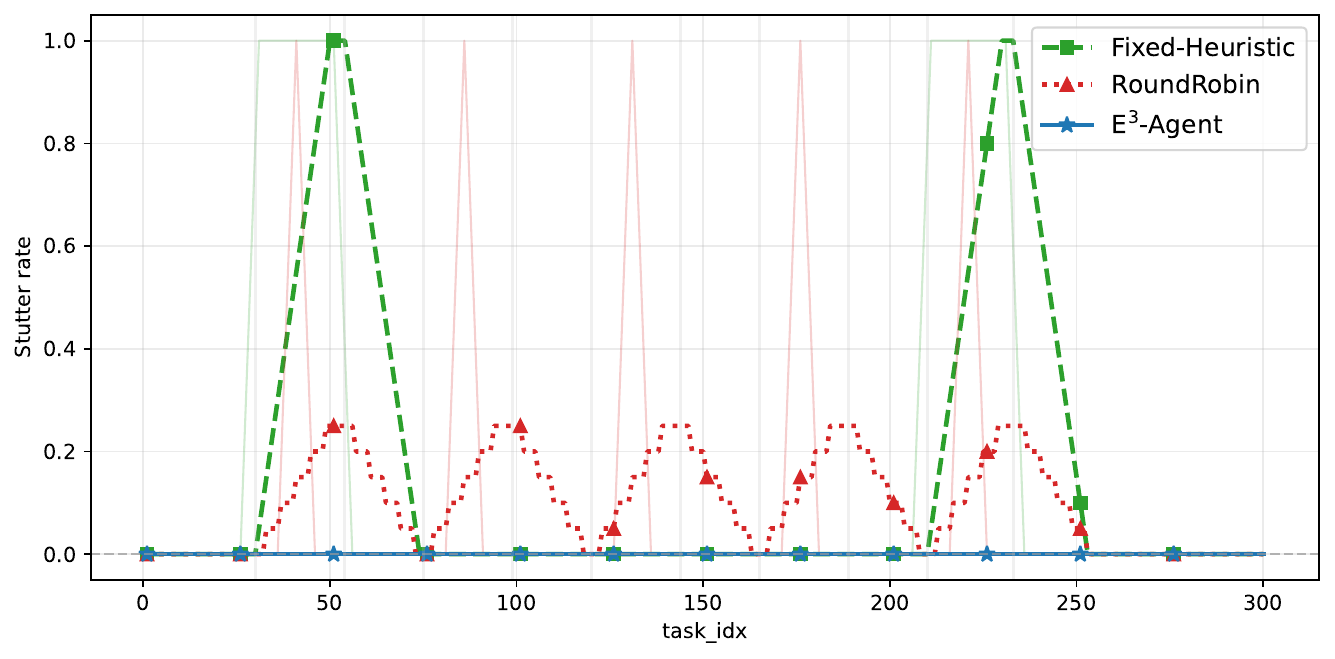}
        \par\small (d) Semantic dynamics (stutter timeline)
    \end{minipage}
    \caption{Dynamic environments: per-task latency trajectories and semantic QoE signal (raw samples every 5 tasks + MA20 for latency curves). MA20 is a 20-task moving average.}
    \label{fig:dyn_2x2}
\end{figure}

\paragraph{Dynamic environments.}
This experiment tests non-stationary adaptation beyond cold start. We evaluate three dynamic regimes, semantic dynamics, device churn dynamics, and hidden drift, each preceded by a short initial warmup period; for this reason, the plotted task index may start after the warmup prefix, for example from $k=50$. Fig.~\ref{fig:dyn_2x2}(a)--(c) shows that E$^{3}$-Agent remains consistently close to Oracle across all regimes, with gaps of only +7.01\% in semantic dynamics, +10.13\% in device churn dynamics, and +8.97\% in hidden drift, as summarized in Table~\ref{tab:main_overall}. In contrast, Fixed-Heuristic and RoundRobin incur gaps exceeding 200\%, highlighting that fixed strategies degrade sharply under hidden and time-varying performance conditions.
In absolute terms, Table~\ref{tab:main_overall} shows that E$^{3}$-Agent reduces average latency from 16--19 s under the static baselines to about 5.3--5.7 s across all three regimes, while tracking the Oracle at about 4.8--5.1 s. This improvement is consistent across distinct sources of non-stationarity. Under device churn dynamics, the feasible device set changes online and queue states reshuffle as devices appear or disappear. Under hidden drift, the service-time mapping shifts without explicit disclosure and must be inferred from residual patterns. The small Oracle gaps indicate that the fast-path router remains effective when driven by continuously updated OPM estimates and risk-aware signals, while the slow-path controller updates calibration and risk overrides only when triggers indicate a regime shift.

Semantic dynamics additionally introduces QoE risk that is not captured by average latency alone. Fig.~\ref{fig:dyn_2x2}(d) reports the stutter timeline under semantic events, and Table~\ref{tab:qoe_overhead} gives the corresponding aggregate stutter and meta-control overhead. E$^{3}$-Agent reduces stutter to 0.00\% in the semantic scenario, while the best static baseline among Fixed-Heuristic and RoundRobin still incurs a 10.00\% stutter rate. This suppression is achieved with 11 LLM invocations and 33 tool calls over 300 tasks, illustrating a practically executable QoE-aware control loop whose control-plane activity is tied to events rather than to every request.

\section{Conclusion and Future Work}
\label{sec:conclusion}
Edge generative inference resource management must increasingly operate under two practical realities: per-device per-model service times are often unknown at deployment time, and they are non-stationary due to semantic events, background load, and device churn. This paper presented E$^{3}$-Agent, an executable and evolving agentic framework that separates a fast-path router from an event-driven LLM meta-controller, enabling online exploration, rapid calibration, and risk-aware mitigation under regime shifts.

We validated E$^{3}$-Agent in a discrete-event simulator driven by MLPerf-derived measurement priors, covering warmup in unknown environments and three dynamic regimes: semantic dynamics, device churn, and hidden drift. E$^{3}$-Agent achieves consistent improvements over static baselines, reduces average latency by about 65\%--73\% in dynamic scenarios while remaining within 7\%--10\% of an online full-information Oracle, and effectively suppresses stutter risk under semantic degradation.

Several limitations point to promising future work. First, deploying E$^{3}$-Agent on real edge devices with production telemetry and actuation interfaces will enable end-to-end validation beyond simulation and reveal additional systems constraints, including monitoring fidelity, delayed or missing feedback, control latency, and safety governance. Second, the current evaluation focuses on compute-side adaptation and does not separately model PHY/MAC channel variability; extending the framework to joint communication-computing control under explicit wireless dynamics is an important next step. Third, expanding workload coverage to more generative tasks and stronger heterogeneity, including text-to-video and multimodal long-sequence generation, larger device pools, and more frequent join or leave, can further stress-test generalization under evolving environments. Finally, incorporating additional system signals and more principled uncertainty estimation, while preserving safety invariants and bounded control-plane overhead, remains an important direction for practical agentic resource management.



\bibliographystyle{scis}
\bibliography{ref}



\end{document}